\documentclass[10pt,journal,compsoc]{IEEEtran}
\ifCLASSOPTIONcompsoc
  \usepackage[nocompress]{cite}
\else
  \usepackage{cite}
\fi
\ifCLASSINFOpdf
\usepackage[pdftex]{graphicx}
\else
\fi
\usepackage{amsmath}
\usepackage{mdwmath}
\usepackage{mdwtab}
\usepackage{eqparbox}
\usepackage{booktabs}

\usepackage{algorithmic}

\usepackage{array}
\newcommand\MYhyperrefoptions{bookmarks=true,bookmarksnumbered=true,
pdfpagemode={UseOutlines},plainpages=false,pdfpagelabels=true,
colorlinks=true,linkcolor={black},citecolor={black},urlcolor={black},
pdftitle={ML Algorithm Synthesizing Domain Knowledge for Fungal Spores Concentration Prediction},
pdfsubject={Typesetting},
pdfauthor={Michael D. Shell},
pdfkeywords={Fungal Conentration Prediction, Machine Learning}}
\ifCLASSINFOpdf
\usepackage[\MYhyperrefoptions,pdftex]{hyperref}
\else
\usepackage[\MYhyperrefoptions,breaklinks=true,dvips]{hyperref}
\usepackage{breakurl}

\fi

\begin{document}
%
\title{ML Algorithm Synthesizing Domain Knowledge for Fungal Spores Concentration Prediction}
%
%
%
%

\author{Md Asif Bin Syed$^{\dagger}$, 
Azmine Toushik Wasi,
and Imtiaz Ahmed
        
\IEEEcompsocitemizethanks{\IEEEcompsocthanksitem M. A. B. Syed and I. Ahmed are with the Department of Industrial and Systems Engineering, West Virginia University, Morgantown, USA.\protect \\
E-mail: ms00110@mix.wvu.edu, imtiaz.ahmed@mail.wvu.edu
\IEEEcompsocthanksitem A. T. Wasi is with the Department of Industrial and Production Engineering, Shahjalal University of Science and Technology, Bangladesh.\protect \\
E-mail: azminetoushik.wasi@gmail.com
\IEEEcompsocthanksitem $^\dagger$Corresponding Author.
\IEEEcompsocthanksitem Code: \href{https://github.com/azminewasi/qcre23-finalist}{\textcolor{blue}{https://github.com/azminewasi/qcre23-finalist}}.
\IEEEcompsocthanksitem Paper Page: \href{https://azminewasi.github.io/research/paper/qcre2023/index.html}{\textcolor{blue}{.../qcre2023}}.
\IEEEcompsocthanksitem This work is selected as \textbf{one of the four finalists of \href{https://www.iise.org/Details.aspx?id=52424}{\textcolor{blue}{ProcessMiner QCRE Data Challenge 2023}}}, and presented on \href{https://www.iise.org/Annual/details.aspx?id=13480}{\textcolor{blue}{IISE Annual Conference and Expo, 2023}}.
\protect\\
}}

%
%

\markboth{\textbf{ML Algorithm Synthesizing Domain Knowledge for Fungal Spores Concentration Prediction}}%
{Shell \MakeLowercase{\textit{et al.}}: Bare Demo of IEEEtran.cls for Computer Society Journals}



\IEEEtitleabstractindextext{%
\begin{abstract}
The pulp and paper manufacturing industry requires precise quality control to ensure pure, contaminant-free end products suitable for various applications. Fungal spore concentration is a crucial metric that affects paper usability, and current testing methods are labor-intensive with delayed results, hindering real-time control strategies. To address this, a machine learning algorithm utilizing time-series data and domain knowledge was proposed. The optimal model employed Ridge Regression achieving an MSE of 2.90 on training and validation data. This approach could lead to significant improvements in efficiency and sustainability by providing real-time predictions for fungal spore concentrations. This paper showcases a promising method for real-time fungal spore concentration prediction, enabling stringent quality control measures in the pulp-and-paper industry. 
\end{abstract}

}


\maketitle

\IEEEdisplaynontitleabstractindextext

%
\IEEEpeerreviewmaketitle

\IEEEraisesectionheading{\section{Introduction}\label{sec:introduction}}

\IEEEPARstart{T}{he} pulp-and-paper manufacturing industry plays a pivotal role in providing essential materials for various applications, including packaging, printing, and writing. Ensuring the production of high-quality, contaminant-free paper products is paramount in this sector. As such, stringent quality control measures are a fundamental aspect of the industry's operations \cite{Bajpai2015}.

Quality control in the pulp-and-paper manufacturing sector involves a comprehensive assessment of various parameters, with a vital focus on the fungal spore concentration. Fungal spores are microscopic particles that can have a detrimental impact on paper quality. When present in excessive amounts, they can lead to a range of issues, including reduced paper strength, increased susceptibility to degradation, and compromised printability. Consequently, addressing fungal spore concentration is essential to meet industry standards and customer expectations.

Current techniques for assessing fungal spore concentration primarily rely on labor-intensive laboratory tests. These methods involve collecting paper samples from the manufacturing process and subjecting them to meticulous analysis, often taking 1-2 days to obtain results. This delay can lead to production inefficiencies and increased costs, highlighting the need for more efficient and real-time monitoring solutions.

Machine learning (ML) has emerged as a promising technology to address these challenges in the pulp-and-paper industry. ML algorithms can be trained to analyze vast amounts of data, including environmental conditions, production parameters, and historical fungal spore data. By processing this information in real-time, ML models can predict fungal spore concentrations within the manufacturing process.

Delay in quality measurement hinders real-time control strategies, emphasizing the need for precise real-time fungal spore concentration predictions to maintain exceptional quality standards. This study showcases the outcome of a data challenge focused on devising a method to predict fungal spore concentration in the pulp-and-paper production process utilizing time-series data. We propose a machine learning algorithm that synthesizes domain knowledge, based on two crucial assumptions. 

\subsection{Our Contributions}
Our contributions are summarized into four folds: 

\begin{itemize}
    \item We design and develop a  novel machine learning-based method utilizing domain knowldge to predict fungal spores concentration effectively considering problem constrains.
    \item Our model requires much less memory than deep learning models because it does not have as many parameters to train. This low memory requirement makes it easier to integrate into embedded systems with limited memory resources.
    \item Our model has a closed-form solution which results in a computationally efficient training process. This means that the training time for Ridge Regression is faster than deep learning techniques like recurrent neural networks (RNNs) or convolutional neural networks (CNNs).
    \item  Our model lightweight and easy to deploy on embedded systems because they do not require large processing power. This makes them ideal for implementing in resource-constrained devices such as embedded systems or Internet of Things (IoT) devices and sensors in the industry.
\end{itemize}

\section{Data Description and Exploratory Data Analysis }
For this study, we were provided with one training set and one testing set \cite{qcre2023}. The training set comprises 1526 observations, containing 113 numerical variables, a single categorical variable, and the target variable, beginning from 2021-01-30 06:23:06. Additionally, the testing data set encompasses 752 observations with an identical variable count, starting from 2021-07-24 06:29:00. To gain an in-depth understanding of the data, we conducted an exploratory data analysis (EDA). The following subsection details the handling of missing values and observed trends for several key variables.
qcre2023

\subsection{{Problem Formulation and Constrains}}
The task at hand involves predicting the target spore concentration for timestamps provided in a dataset. Notably, the timestamps in the test set may not necessarily occur after those in the training data, introducing an element of temporal unpredictability. A critical constraint in this problem lies in the temporal alignment: when predicting the target value for a given test set timestamp (t1), any training data with timestamps beyond t1 is expressly prohibited from being used \cite{qcre2023}. This constraint reflects the real-world scenario where predictions must rely solely on historical information up to the prediction point, ensuring that the model's forecasting capabilities align with chronological precedence. This challenge falls within the realm of time-series forecasting, where the model's effectiveness in capturing temporal patterns and dependencies is pivotal to its predictive accuracy.

\subsection{{Missing Values}}
In examining our training data, we assessed the presence of missing values, identifying several variables with missing data, as outlined in the Fig 1. We used most frequent value imputation.

\begin{figure}[h!] 
\centering {\includegraphics[scale=0.55]{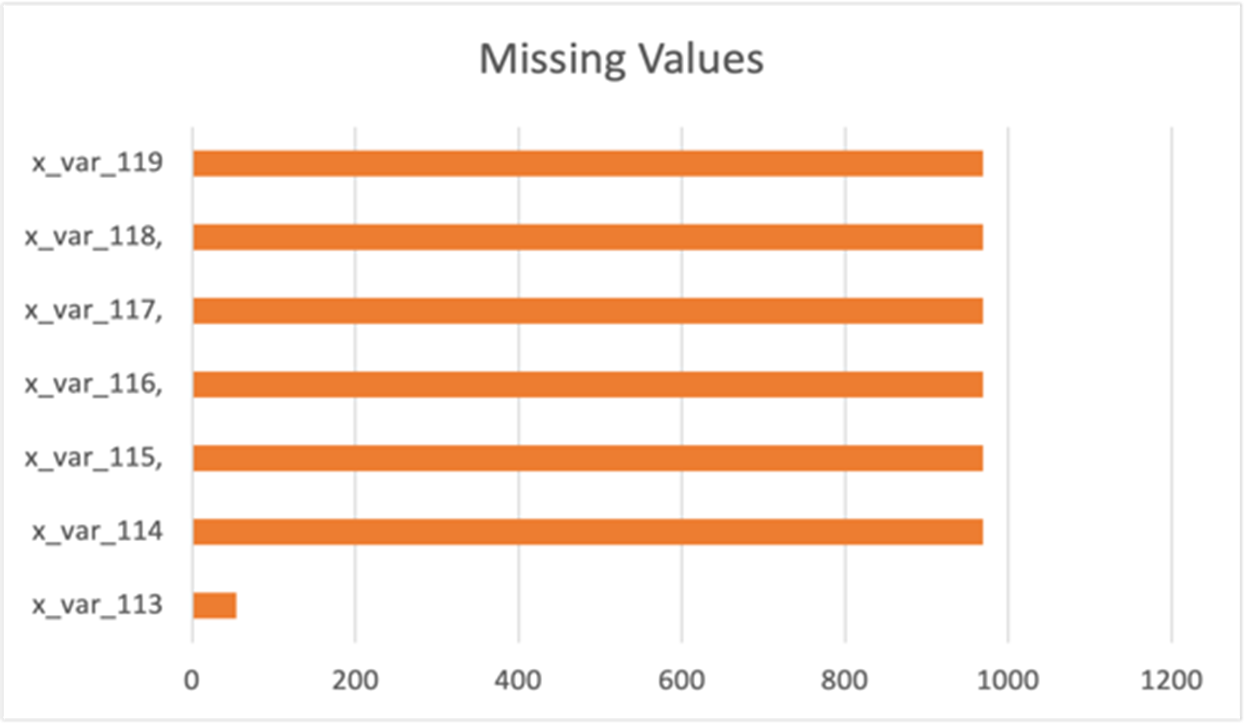}}
\caption{Missing Values in the dataset.}\label{fig:missing-values}
\end{figure}

\begin{table*}[pht]
\centering
\caption{Impact of Hyperparameter $\alpha$ on Performance Metrics}
    \begin{tabular}{l c c c c c c c c}
        \toprule
        $\alpha$ & 0.05 & 0.1 & 0.5 & 0.8 & 1.5 & 2 & 2.5 & 3 \\
        \midrule
        \textbf{MAE} Value & 0.51 & 0.51 & 0.51 & 0.51 & 0.48 & 0.43 & 0.52 & 0.6 \\
        \textbf{MSE} Value & 2.9 & 2.9 & 2.9 & 2.9 & 2.74 & 2.53 & 2.94 & 3.35 \\
        \textbf{RMSE} Value & 1.7 & 1.7 & 1.7 & 1.7 & 1.65 & 1.59 & 1.72 & 1.83 \\
        \textbf{R-squared} Value & 0.98 & 0.98 & 0.98 & 0.98 & 0.98 & 0.98 & 0.98 & 0.98 \\
        \bottomrule
    \end{tabular}
\end{table*}

\subsection{{Analysis of Time dependency of variables}}
Besides missing values, we investigated the potential dependency on time. By observing the provided figure, we attempted to discern any time dependency among the features. As depicted in Fig. 2, we found no substantial time dependency for the majority of variables, with a few exceptions (e.g., variable 41).

\begin{figure}[ht]
\centering
\includegraphics[width=8cm]{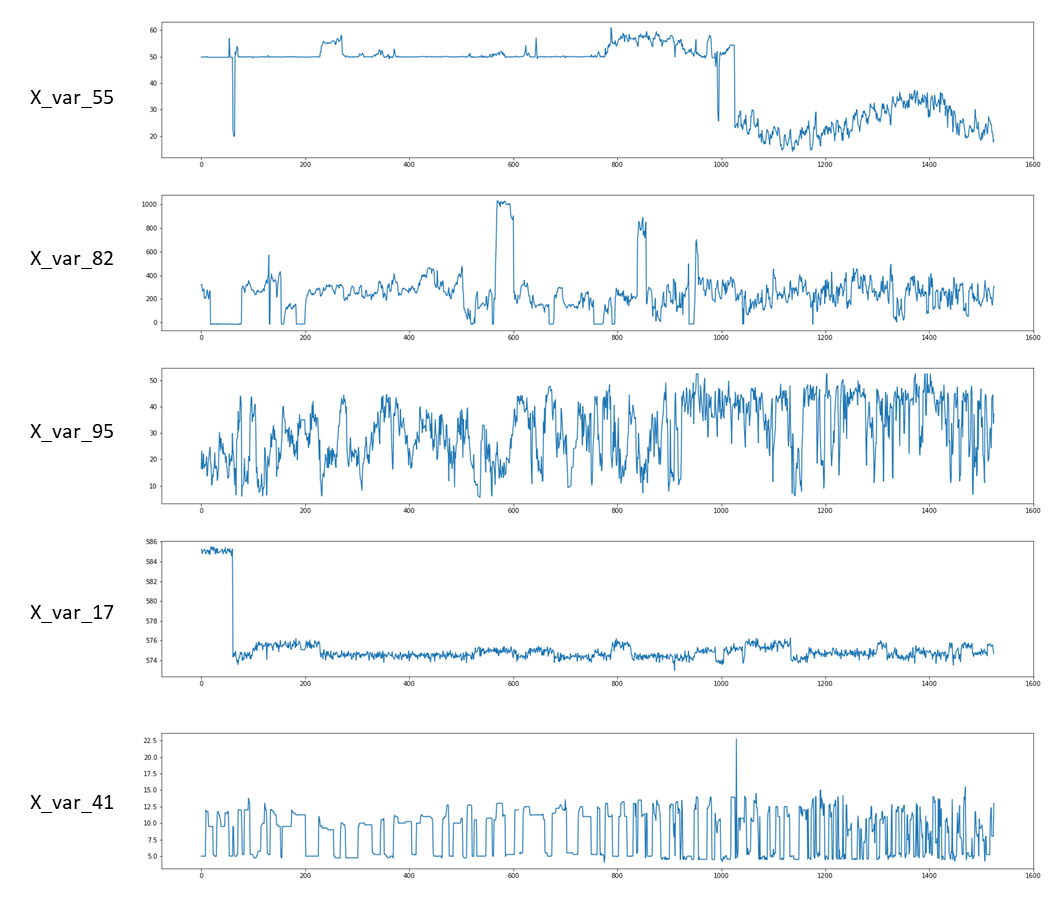}
\caption{Analysis of Time dependency of variables}
\end{figure}

\section{Methodology} 
Upon reviewing the exploratory data analysis ad considering the project constraint that prohibits training data beyond t1 when predicting t1, we develop the following algorithm depicted in the flow chart shown in Fig 3.  The following subsection discusses the components of the algorithm. 
\begin{figure}[ht]
\centering
\includegraphics[width=8cm]{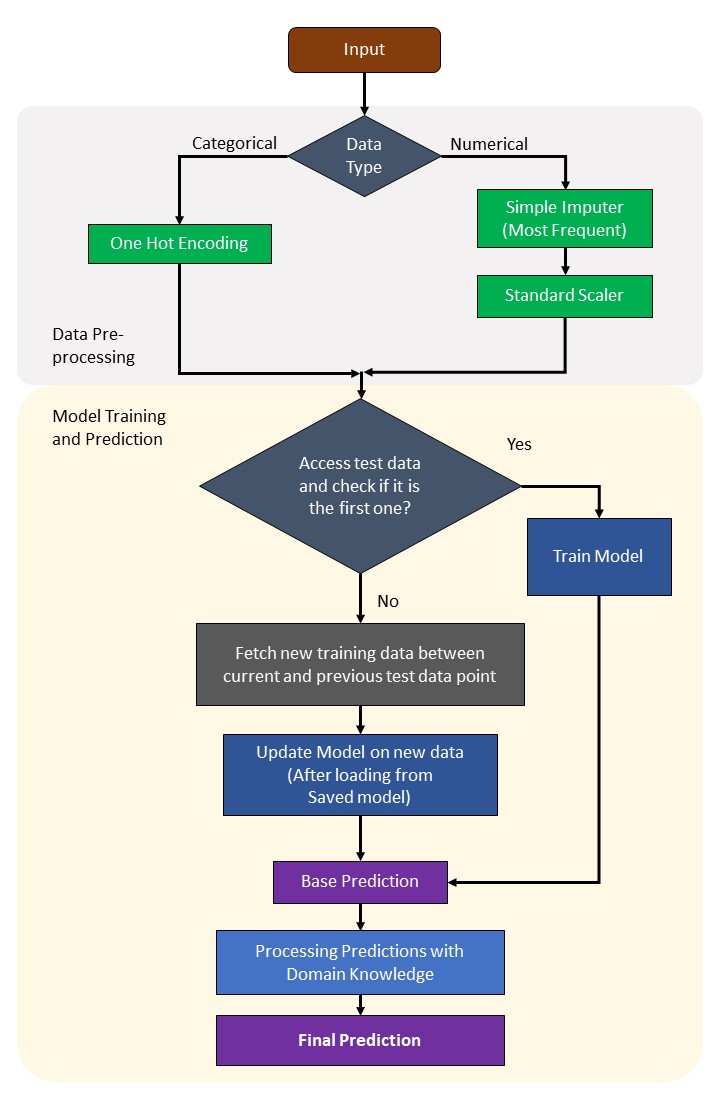}
\caption{Model Architecture} 
\end{figure}

\subsection{Train-test Splitting}
To evaluate model performance, we devised a distinctive data partitioning strategy. Initially, the first 40\% of the dataset was allocated for training, ensuring sufficient data availability. Subsequently, the remaining 60\% was divided into testing and training subsets, which were merged with the previous partition. This approach emulates the original problem, providing 750 initial data points for model training.

\begin{figure}[ht]
\centering
\includegraphics[width=7cm]{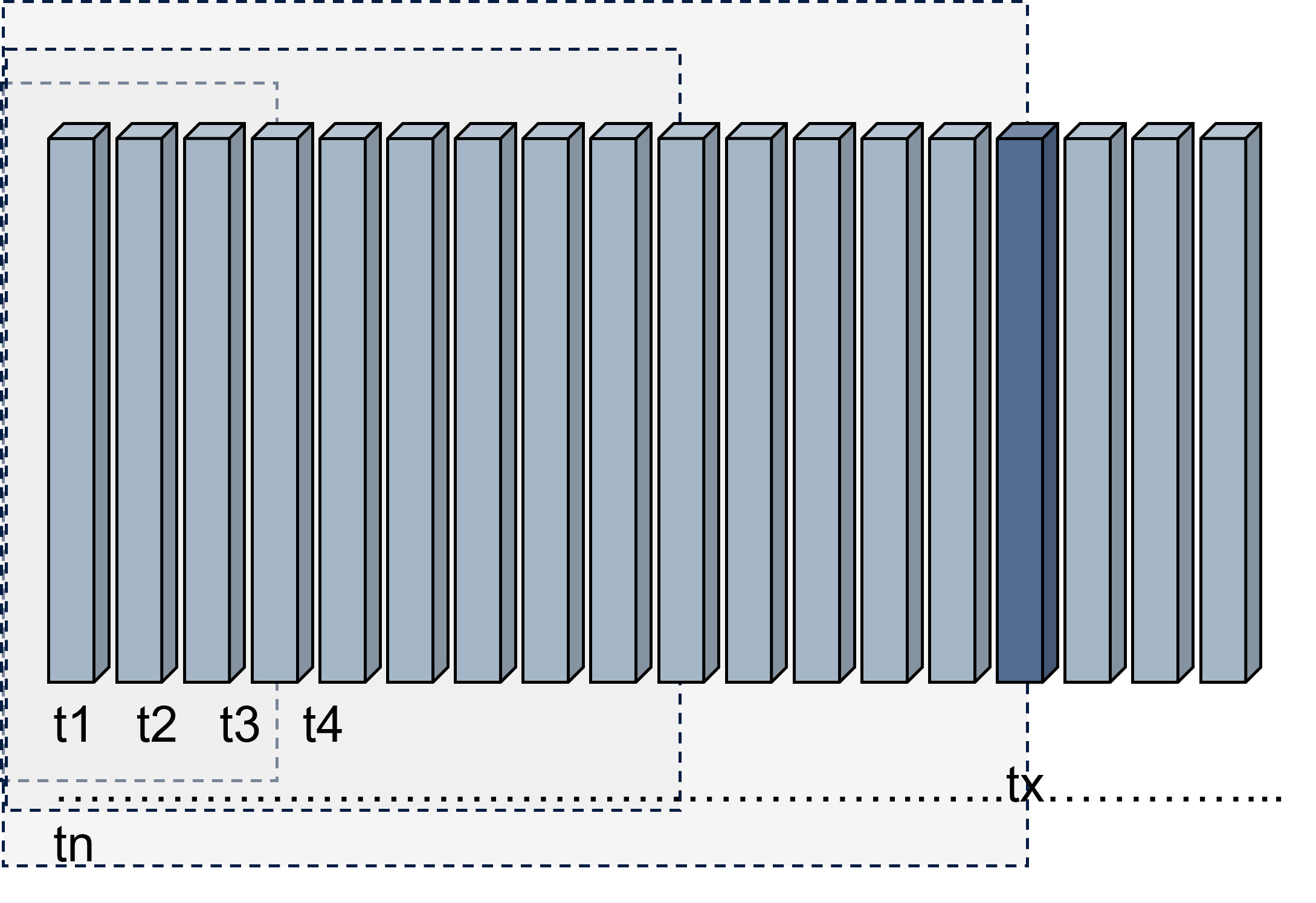}
\caption{Training Data Selection based on Problem Constrains}
\end{figure}
\subsection{Model Architecture} 

\begin{figure}[ht]
\centering
\includegraphics[width=8cm]{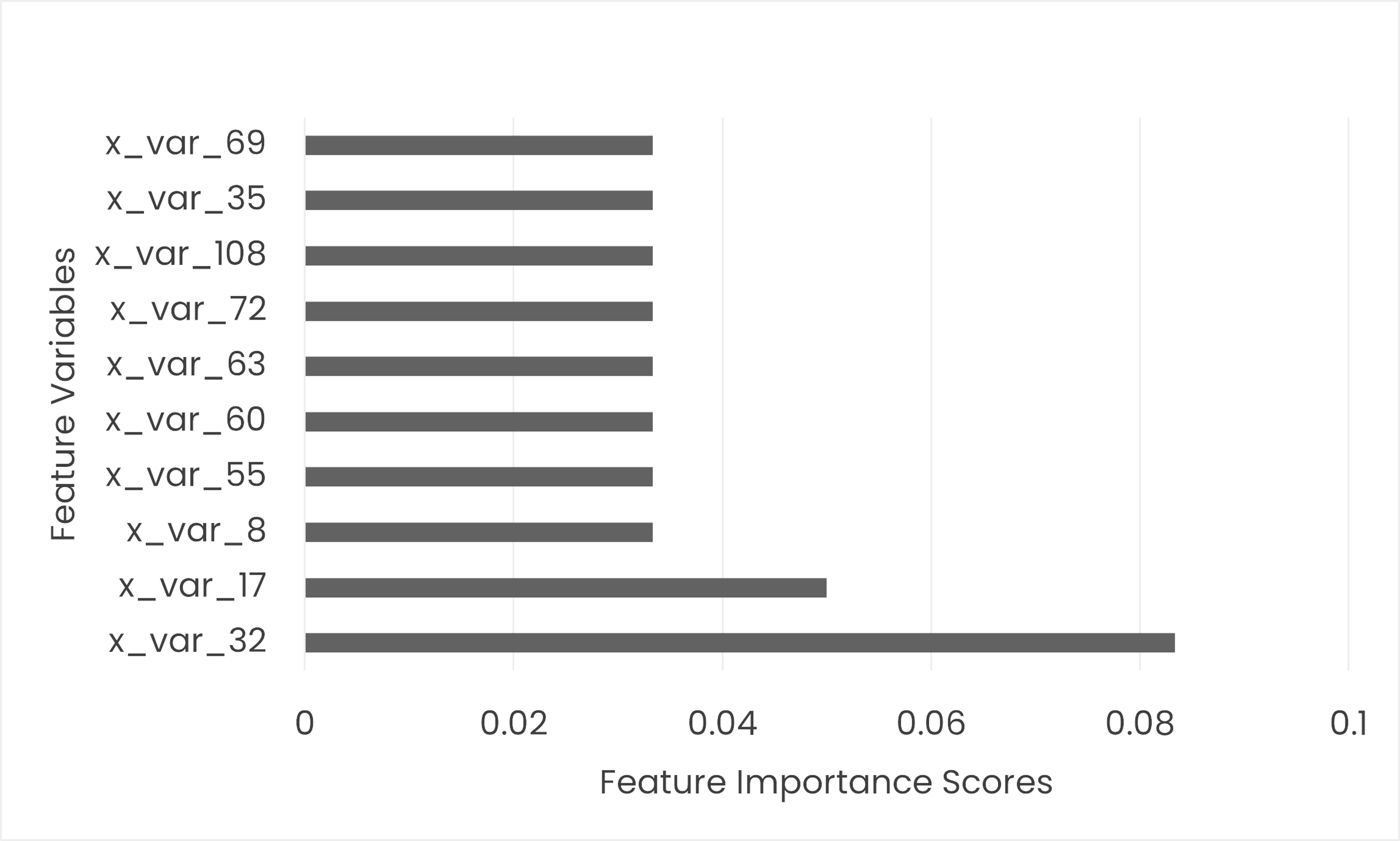}
\caption{Top features by a Random Forest Model} \label{fig:Fi-rf}
\end{figure}

To mitigate noise, feature selection techniques were employed, including Principal Component Analysis, Random Forest Regressor, and SelectKBest, enabling identification of the most relevant features and enhancing model validity. One example is shown in Figure \ref{fig:Fi-rf}.

\begin{figure}[ht]
\centering
\includegraphics[width=8cm]{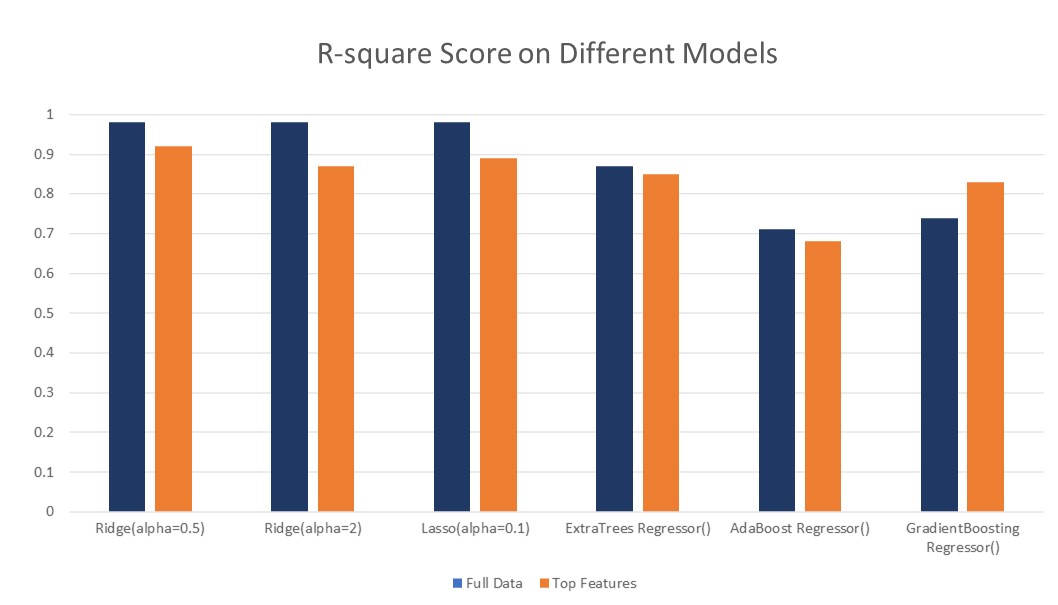}
\caption{R2 Score Comparison} \label{fig:Fi-r2}
\end{figure}

\begin{figure}[ht]
\centering
\includegraphics[width=8cm]{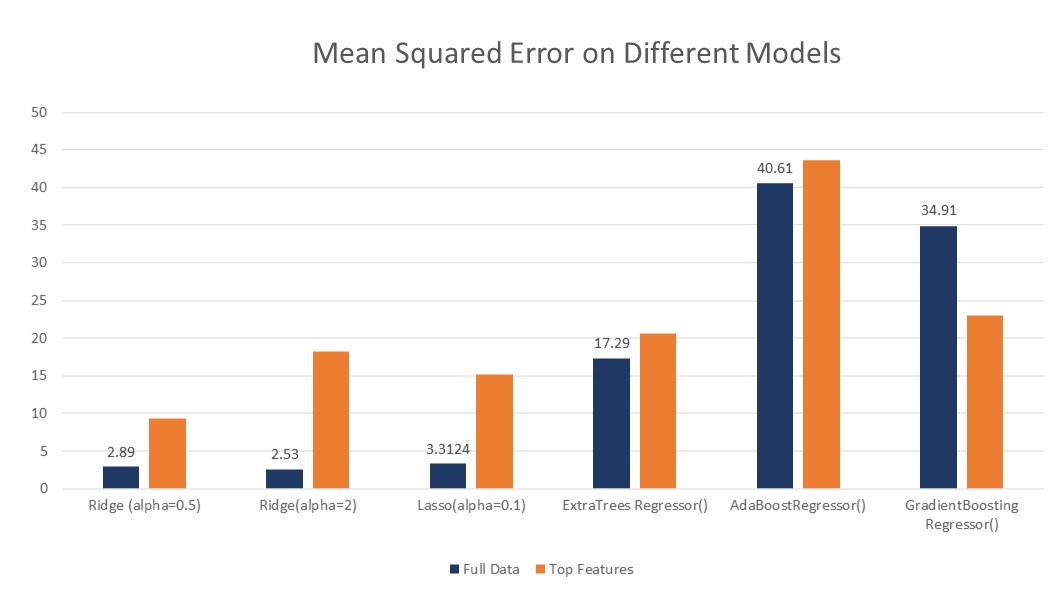}
\caption{Mean Squarred Error Score Comparison} \label{fig:Fi-mse}
\end{figure}

But, Figure \ref{fig:Fi-r2} and \ref{fig:Fi-mse} suggests that the model scores better with all features, rather than top features.

\subsection{Model Selection and Training} 
We conducted experiments with prevalent machine learning algorithms, including Linear, Ridge, and Lasso regression, Random Forest, Extreme Gradient Boosting (XGBoost), and Adaptive Boosting. A constraint required training only on data up to t for predicting t+1. To accommodate this, model parameters were stored in a binary serialization file, enabling efficient parameter updates without retraining on the entire dataset. 

We found that Ridge Regression\cite{bhlitem137258} performed exceptionally well on our sensor dataset with a large number of input columns. This is likely because Ridge Regression's regularization technique helps to reduce overfitting and handle multicollinearity, leading to stable coefficient estimates and better predictive performance on new data. it is a regularization technique that helps to reduce overfitting in ML models. It adds a penalty term to the cost function which reduces the magnitude of the coefficients, thus limiting the model's complexity and making it less prone to overfitting. It performs well when there is multicollinearity between the independent variables. Multicollinearity occurs when two or more independent variables are highly correlated with each other, which can lead to unstable and unreliable coefficient estimates in traditional linear regression. The penalty term in Ridge Regression helps to stabilize these estimates by reducing their variance. It can provide a solution even when the data is inconsistent or when the number of samples is smaller than the number of features. This is because it introduces bias into the estimates, which can help to overcome problems caused by inconsistencies in the data.

\subsection{Prediction Synthesizing the Domain Knowledge} 

\begin{figure}[ht]
\centering
\includegraphics[width=8cm]{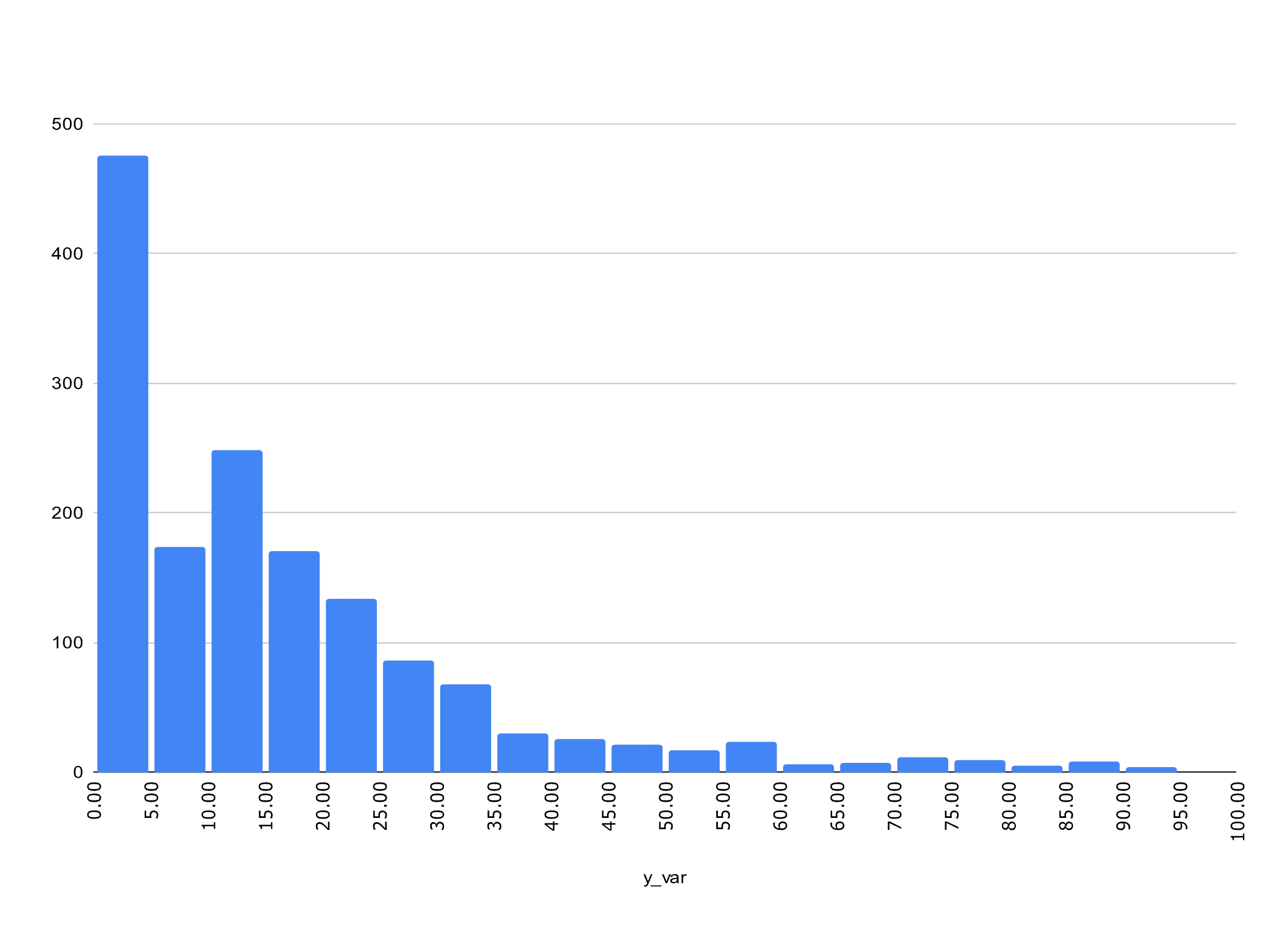}
\caption{Target Variable Intervals} \label{fig:tv}
\end{figure}
From Figure \ref{fig:tv}, we can notice that target variable seems to be an integer which is divisible by 5. Our initial predictions are a lot close to multiples of 5,  like we have 11, 11.5, 10.5 more than 12-12.5 which is a more uncertain stage. From this, we came to two assumptions which we are referring as domain knowledge. The two assumptions are: 
\begin{enumerate}
  \item All the "y\_var" are the multipliers of the 5. Our assumption is the measurement scale has the precision of 5. 
  
  \item As the concentration cannot be negative so we have chosen any negative prediction as zero. 
\end{enumerate}

The equation is:

\begin{equation}
f(x)=\left\{\begin{array}{lr}
0,  \text { if } x \leq 2.5, \\
\hspace{6mm}\text { or } x \text { is an integer multiple of } 5,\\
\hspace{6mm}\text { or } x<0 \\

5 * \text { floor }(x / 5)+5,  \\
\hspace{6mm}\text { if }(x-0.01) \text { is an integer multiple of } 5 \\
5^* \text { floor }(x / 5),  \text { otherwise }
\end{array}\right\}
\end{equation}

\section{Results} 
Upon experimenting with training data, partitioned into training and validation sets that emulate the original testing dataset, we observed the following results. Using all available variables as input, alongside selected features via algorithms such as Random Forest, PCA, and SelectKBest, Random Forest demonstrated superior performance in terms of MSE and MAE. Notably, the cleaned and preprocessed original dataset outperformed feature-selected versions. A comparative analysis between the full data and Random Forest models, with respect to MSE and MAE, is provided.

\subsection{Ablation Study}
Ridge regression was selected due to its superior performance compared to other algorithms, including Linear Regression, Adaptive Boosting, XGBoost, and Extra Trees Regressor. An optimal hyperparameter tuning was achieved with an alpha value of 2, as it demonstrated enhanced performance relative to alternatives (as shown in table \textbf{\ref{tab:split-metrics}}).

\begin{table}[ht]
    \centering
    \caption{Cross Validation Scores}
    \vspace{2mm}
    \label{tab:split-metrics}
    \begin{tabular}{c c c c c c }
        \toprule
        Split & MAE & MSE & RMSE & R-squared \\
        \midrule
        1 & 1.04 & 7.29 & 2.7 & 0.98 \\
        2 & 0.21 & 1.04 & 1.02 & 0.98 \\
        3 & 0.62 & 3.12 & 1.77 & 0.96 \\
        4 & 0.33 & 1.42 & 1.19 & 0.99 \\
        \bottomrule
    \end{tabular}
\end{table}

To mitigate overfitting, cross-validation was employed. The outcomes obtained with k=5 are presented, revealing satisfactory results.

\section{Conclusion}\label{s:conclusion}
In this study, we introduce an algorithm to predict fungal spore concentration, encompassing two components: a traditional machine learning approach for real-time training, and domain knowledge synthesis with predictions. Our chosen method effectively balances the bias-variance tradeoff, as demonstrated by strong training results reflecting the problem's intrinsic model. Future research incorporating controlled deep learning approaches may yield even more accurate solutions. 

{\small
\bibliographystyle{ieee_fullname}
\bibliography{main}
}

\end{document}